\documentclass{ifacconf}

\input{packages}

\newcommand{\be}{\begin{equation}}
\newcommand{\ee}{\end{equation}}
\newcommand{\ba}{\begin{array}}
\newcommand{\ea}{\end{array}}
\newcommand{\bea}{\begin{eqnarray}}
\newcommand{\eea}{\end{eqnarray}}

\newcommand{\eig}{{\mathrm{eig}}}       
\newcommand{\tran}{^{\mbox{\scriptsize T}}}  

\newcommand{\vbar}{\raisebox{.17ex}{\rule{.04em}{1.35ex}}}
\newcommand{\vbarind}{\raisebox{.01ex}{\rule{.04em}{1.1ex}}}
\newcommand{\D}{\ifmmode {\rm I}\hspace{-.2em}{\rm D} \else ${\rm I}\hspace{-.2em}{\rm D}$ \fi}
\newcommand{\T}{\ifmmode {\rm I}\hspace{-.2em}{\rm T} \else ${\rm I}\hspace{-.2em}{\rm T}$ \fi}
\newcommand{\B}{\ifmmode {\rm I}\hspace{-.2em}{\rm B} \else \mbox{${\rm I}\hspace{-.2em}{\rm B}$} \fi}
\newcommand{\Hil}{\ifmmode {\rm I}\hspace{-.2em}{\rm H} \else \mbox{${\rm I}\hspace{-.2em}{\rm H}$} \fi}
\newcommand{\C}{\ifmmode \hspace{.2em}\vbar\hspace{-.31em}{\rm C} \else \mbox{$\hspace{.2em}\vbar\hspace{-.31em}{\rm C}$} \fi}
\newcommand{\Cind}{\ifmmode \hspace{.2em}\vbarind\hspace{-.25em}{\rm C} \else \mbox{$\hspace{.2em}\vbarind\hspace{-.25em}{\rm C}$} \fi}
\newcommand{\Q}{\ifmmode \hspace{.2em}\vbar\hspace{-.31em}{\rm Q} \else \mbox{$\hspace{.2em}\vbar\hspace{-.31em}{\rm Q}$} \fi}
\newcommand{\Z}{\ifmmode {\rm Z}\hspace{-.28em}{\rm Z} \else ${\rm Z}\hspace{-.38em}{\rm Z}$ \fi}

\renewcommand{\vec}[1]{{\bf{#1}}}     

\newcommand{\bc}{\mbox {\boldmath $c$}}

\newcommand{\bg}{\mbox {\boldmath $g$}}

\newcommand{\br}{\mbox {\boldmath $r$}}

\newcommand{\bu}{\mbox {\boldmath $u$}}
\newcommand{\bv}{\mbox {\boldmath $v$}}
\newcommand{\bw}{\mbox {\boldmath $w$}}
\newcommand{\bx}{\mbox {\boldmath $x$}}
\newcommand{\bX}{\mbox {\boldmath $X$}}

\newcommand{\bW}{\mbox {\boldmath $W$}}
\newcommand{\by}{\mbox {\boldmath $y$}}
\newcommand{\bz}{\mbox {\boldmath $z$}}

\newcommand{\calR}{\mathcal{R}}

\newcommand{\calW}{\mathcal{W}}

\newcommand{\R}{\mathbb{R}}
\newcommand{\N}{\mathbb{N}}

  \maketitle

\newcommand{\Proj}{\texttt{Proj}}

\newcommand{\Conv}{\mathrm{Conv}}

\begin{document}

\begin{frontmatter}

\title{Communication-efficient Variance-reduced Stochastic Gradient Descent}
\author{Hossein S. Ghadikolaei and Sindri Magn\'usson}

\address{School of EECS, KTH Royal Institute of Technology, Stockholm, Sweden (emails: \{hshokri, sindrim\}@kth.se).}

\begin{abstract}
We consider the problem of communication efficient distributed optimization where multiple nodes exchange important algorithm information in every iteration to solve large problems. In particular, we focus on the stochastic variance-reduced gradient and propose a novel approach to make it communication-efficient. That is, we compress the communicated information to a few bits while preserving the linear convergence rate of the original uncompressed algorithm. Comprehensive theoretical and numerical analyses on real datasets reveal that our algorithm can significantly reduce the communication complexity, by as much as 95\%, with almost no noticeable penalty. Moreover, it is much more robust to quantization (in terms of maintaining the true minimizer and the convergence rate) than the state-of-the-art algorithms for solving distributed optimization problems. Our results have important implications for using machine learning over internet-of-things and mobile networks.
\end{abstract}


\end{frontmatter}

\section{Introduction}
The recent success of artificial intelligence (AI) and large-scale machine learning is mainly due to the availability of big datasets and large platforms that can provide vast amounts of computational and communication resources for training. However, there are increasing demands to extend large-scale machine learning to a general networked AI where communication and energy resources may be rare and expensive commodities. Use cases include machine leaning over internet-of-things, edge computing, and vehicular networks~\citep{li2018learning,Jiang2019LowLatency}.

We address the problem of computational and communication-efficient distributed optimization. Consider optimizing a finite sum of differentiable functions $\{f_{i}: \R^d \mapsto \R\}_{i \in [N]}$ with corresponding gradients $\{\bg_{i}: \R^d \mapsto \R^d\}_{i \in [N]}$. Iteration $k$ of a common solution algorithm involves finding the gradients $\bg_i(\bw_k)$ at parameter $\bw_k$, sending them back to a central controller (master node), updating the parameters to $\bw_{k+1}$ at the master node, and broadcasting it to start the next iteration. In a distributed optimization setting, one can divide data samples among $N$ computational (worker) nodes for parallel computing. This approach improves privacy and obtains a good tradeoff between computational complexity and convergence rate, thanks to new algorithms like stochastic variance reduced gradient (SVRG) and its variants~\citep{johnson2013accelerating,konevcny2016federated}. However, it raises a serious challenge of communication complexity, defined as the level of information exchange among the workers and the master node to ensure the convergence of the iterative solution algorithm (like SVRG)~\citep{Bottou2018SIAM}. It has been shown in many applications with geographically-separated workers that the communication complexity dominates the run-time of a distributed optimization \citep{konevcny2016federated}. This becomes particularly important when we implement machine learning and distributed optimization algorithms on bandwidth and battery-limited wireless networks. In that case, the latency (in remote industrial operation) or energy consumption (in low power internet-of-things) may render the ultimate solution and consequently the distributed algorithm useless.

There is a recent wave of attempts to address the communication complexity of distributed optimization. A prominent approach is to use a lossy-compression (realized through quantization with few bits) of the parameter and gradient vectors in every iteration to save the communication resources. This approach has been applied to both deterministic~\citep{jordan2018communication,magnusson2017convergence,magnusson2019maintaining} and stochastic~\citep{bernstein2018signsgd,wen2017terngrad,de2018high} algorithms.
However, the conventional wisdom, validated by empirical observations, is that there exists a precision-accuracy tradeoff: the fewer the number of bits per iteration the lower the accuracy of the final solution. Some recent studies, however, challenged that wisdom in the deterministic~\citep{magnusson2019maintaining} and stochastic~\citep{de2018high} settings, and showed that proper quantization approaches can maintain the convergence (both to the true minimizer and the convergence rate). Using similar ideas to \citep{magnusson2019maintaining}, but in the stochastic gradient setting, we propose a novel adaptive quantization approach that theoretically ensures a linear convergence rate of a pseudo-contractive algorithm.
The closest work to us is \citep{de2018high} where the authors proposed an adaptive quantization scheme to iterate based on low-precision $\bw_k$ while maintaining high-accuracy training performance. Our novel algorithm includes low-precision gradients (in uplink transmission), in addition to the low-precision parameters (in downlink transmission), into the iterations of SVRG. This is important in many real applications as the uplink channel may have a much lower speed than the downlink channel~\citep{furht2016long} in which sending low-precision gradients leads to a substantial performance gain. Handling this modification involves a non-trivial modification of the original SVRG algorithm using a memory unit, as described in Section~\ref{sec: MainResults}.

In this paper, we focus on SVRG for the class of strongly convex and smooth functions (which appear in many applications including channel estimation, localization, resource allocation, and network optimization~\citep{convex_boyd}). Our proposed algorithm can recover the minimizer while maintaining the linear convergence rate with only a fixed number of bits for information exchange among the workers and master node. This is possible due to the fact that, with minor modification of the SVRG algorithm, the norm of the gradient and therefore the difference between $\bw_{k} - \bw_{k-1}$ gets smaller for larger $k$. Consequently, by properly tuning hyper-parameters of the quantization at every step, we can increase the quantization resolution without adding any additional bits.
Comprehensive theoretical and numerical analyses on real datasets reveal that our algorithm can significantly reduce the communication complexity, by as much as 95\%, with almost no noticeable penalty (neither in training nor in test). Our results also indicate that our quantized SVRG approach is much more robust to quantization (in terms of maintaining the true minimizer and the convergence rate) than other state-of-the-art algorithms for solving distributed optimization problems.

The rest of the paper is organized as follows. Section~\ref{sec: BasicAssumptions} presents the problem setting and basic assumptions. Section~\ref{sec: MainResults} provides our main theoretical results. We apply our algorithms on real datasets in Section~\ref{sec: experiments}, and then conclude the paper in Section~\ref{sec: conclusions}. Due to a lack of space, we have moved all the proofs in the extended version of this paper~\citep{Ghadikolaei2019communicationSVRG}.

\emph{Notation:}
Normal font $w$ or $W$, bold font small-case $\bw$, bold-font capital letter $\bW$, and calligraphic font $\calW$ denote scalar, vector, matrix, and set, respectively. We let $[N] = \{1,2,\ldots,N\}$ for any integer $N$. We denote by $\|\cdot\|$ the $l_2$ norm, by $\Conv(\calW)$ the convex hull of set $\calW$, by $\Proj(\bw,\calW)$ the projection of vector $\bw$ onto set $\calW$, by $[\bw]_i$ the $i$-th entry of vector $\bw$, and by $\bw{\tran}$ the transpose of $\bw$.

\section{System Model}\label{sec: BasicAssumptions}
\subsection{Problem Statement and Algorithm Model}
Consider a network of $N$ computational nodes (workers) that cooperatively solve a distributed computational problem involving an objective function $f(\bw)$. Let tuple $(\bx_{ij}, \by_{ij})$ denote data sample $j$ of worker $i \in [N]$, and $\bw$ denote the shared model parameter with dimension $d$. Our optimization problem is
\begin{align}\label{eq: our-optimization}
\bw^\star \in \min_{\bw\in\R^d} f(\bw) & := \frac{1}{N}\sum_{i \in [N]} \underbrace{\frac{1}{N_i}\sum_{j \in [N_i]} f(\bw; \bx_{ij}, \by_{ij})}_{f_i(\bw)} \:,
\end{align}
where $\bg_i$ and $\bg$ denote the gradient of $f_i$ and $f$, respectively, and $N_i$ is the number of samples at node $i \in [N]$.

SVRG is among the state-of-the-art algorithms to balance computational loads of obtaining the full gradients $\{\bg_i\}_i$ and the convergence rate of the distributed algorithm~\citep{johnson2013accelerating}. The inner loop of SVRG (also called epoch) consists of $T$ iterations, wherein the parameters will be sequentially updated using the gradient of $f_{\xi}$ for a randomly chosen index $\xi$. In particular, in the beginning, the master node broadcasts parameter vector $\widetilde{\bw}_{k}$. It then broadcasts $\bw_{k,t-1}$, realizes random variable $\xi \in [N]$, and receives $g_\xi (\bw_{k,t-1})$, in every inner-loop iteration $t$. At the end of the inner loop, the master node updates $\widetilde{\bg}_k$ and $\widetilde{\bw}_{k}$ for the next epoch, see Algorithm~\ref{alg:QSVRG1}.

For the sake of mathematical analysis, we limit the class of objective functions to strongly convex and smooth function, though our approach is applicable to invex~\citep{karimi2016linear} and multi-convex~\citep{xu2013block} structures (like a deep neural network training optimization problem) after some minor modifications.
\begin{assum}\label{assumption: mubeta}
We assume that $f(\bw)$ is $\mu$-strongly convex and its gradient terms $\{\bg_i\}_i$ are $L$-Lipschitz $\forall i \in [N]$, i.e.,
\begin{subequations}
\begin{align}
\label{eq:StronglyConvex}
\left(\bw - \by\right)^T \left(\bg(\bw) - \bg(\by) \right) & \geq \mu \|\bw - \by\|^2 \:, \\
\label{eq:Lipschitzness}
\|\bg_i(\bw) - \bg_i(\by) \| & \leq L \|\bw - \by\| \:.
\end{align}
\end{subequations}
\end{assum}
Next, we define the quantization space and quantization operator.

\subsection{Quantization}
As stated, the SVRG iterations require an exchange of the current parameters and gradients among the workers and the master node. In practice (e.g., patrolling and surveillance, and smart manufacturing), the nodes must first represent real-valued vectors ($\bw_k$ and $\{\bg_i\}_{i\in[N]}$) by a finite number of bits through the so-called quantization/compression operation. The current distributed optimization systems usually use single/double-precision floating-point (32/64 bits mapped according to the IEEE 754 format) to show every entry of those vectors and assume no gap between the quantized values and the original value (infinite number of bits assumption). In this work, we show that a very simple alternative
mapping with much fewer bits can be used to maintain the convergence properties of unquantized SVRG.

\begin{defn}[Quantization Space and Quantizer]\label{def: Quantization-space}
Let \newline $\calR := \calR(\bc, \br, \{b_i\}_{i\in[d]})\subseteq \R^{d}$ for some positive vector $\br$ be a $d$-dimensional lattice of $2^b$ points (quantization space) centered at $\bc \in \R^{d}$ with $2^{b_i}$ points in coordinate $i \in [d]$ (where $b_i\in \N^+$) such that $\sum_{i \in [d]} b_i = b$. For any vector $\bw$, we define quantizer $q\colon \bw \mapsto \Proj(\bw,\calR)$. For notation simplicity, we assume that $b/d$ is an integer.
\end{defn}
Fig.~\ref{fig:Quant} illustrates Definition~\ref{def: Quantization-space}, and Algorithm~\ref{alg:QSVRG1} shows the iterations of quantized SVRG assuming that $\calR_{w,k}$ and $\calR_{g,k}$ denote quantization grids for parameter $\bw$ and for gradient $\bg$, respectively. Various definitions of the projection operator characterize various quantizers. For any point $\bw \in \Conv(\calR)$, let $\{\bv_1,\bv_2,\ldots,\bv_{2^d}\}$ be the vertices of a $d$-dimensional cube in $\calR$ that contains $\bw$. Next, we define unbiased random quantizer (URQ) as
follows:
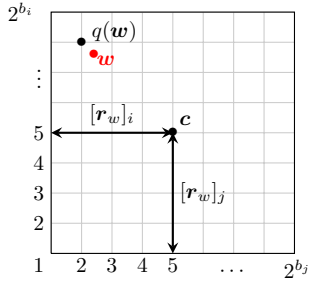
\begin{figure}
 \centering
 {\begin{tikzpicture}[scale=0.8, every node/.style={transform shape}]
\draw[step=0.5cm,gray!40,thin] (-2,-2) grid (2,2);
\draw[thin] (-2,-2) -- (2,-2) -- (2,2) -- (-2,2) -- (-2,-2);
\draw[thick,<->,>=stealth] (-2,0) -- (0,0); 
\draw[thick,<->,>=stealth] (0,-2) -- (0,0);
\node[] at (0,0) {$\bullet$};
\node[] at (-1.3,1.3) {$\color{red}{\bullet}$};
\node[] at (-1.1,1.2) {$\color{red}{\bw}$};
\node[] at (-1.5,1.5) {$\bullet$};
\node[] at (-0.96,1.67) {$q(\bw)$};
\node[] at (0.2,0.2) {$\bc$};
\node[] at (0.5,-1) {$[\br_w]_j$};
\node[] at (-1,0.3) {$[\br_w]_i$};



\node[] at (-2.2,-2.2) {$1$};
\node[] at (-2.2,-1.5) {$2$};
\node[] at (-2.2,-1) {$3$};
\node[] at (-2.2,-0.5) {$4$};
\node[] at (-2.2,0) {$5$};
\node[] at (-2.2,1) {$\vdots$};
\node[] at (-2.5,2) {$2^{b_i} $};

\node[] at (-1.5,-2.2) {$2$};
\node[] at (-1,-2.2) {$3$};
\node[] at (-0.5,-2.2) {$4$};
\node[] at (0,-2.2) {$5$};
\node[] at (1,-2.3) {$\cdots$};
\node[] at (2.05,-2.25) {$2^{b_j}$};

\end{tikzpicture}}

 \caption{Our quantization space $\calR$, determined by a grid center $\bc$, coverage radius in every coordinate $\{[\br_{w}]_i\}_{i \in [d]}$, and number of bits $\{b_i\}_{i \in [d]}$. The quantizer maps $\bw \in \Conv(\calR)$ to one of its coordinates using mapping $q$.}
 \label{fig:Quant}
\end{figure}

\begin{algorithm}[t]
\caption{Quantized SVRG}
\label{alg:QSVRG1}
\begin{algorithmic}[1]
\State \textbf{Inputs:} Epoch length $T$, number of epochs $K$, $N$, step size sequence $(\alpha_k)_k$, number of bits
\For{$k =1,2,\ldots,K$}
\State $\widetilde{\bg}_k \gets \frac{1}{N}\sum_{i\in[N]} \bg_i\left(\widetilde{\bw}_k\right)$
\State Set ${\calR}_{w,k}$ and $\{{\calR}_{\xi,k}\}_{\xi \in [N]}$
\State $\bw_{k,0} \gets \widetilde{\bw}_k$
\For{$t=1,2,\ldots,T$}
\State Sample $\xi$ uniformly from $[N]$
\State Send $\bg_\xi \left(\bw_{k,t-1}\right)$ and $q\left(\bg_\xi\left(\widetilde{\bw}_{k}\right) ; {\calR}_{\xi,k}\right)$
\State $\small{\vec{u}_{k,t} \gets \bw_{k,t-1} - \alpha_k \left(\bg_\xi (\bw_{k,t-1}) - q\left(\bg_\xi(\widetilde{\bw}_{k}) \right) + \widetilde{\bg}_k  \right)}$
\State $\bw_{k,t} \gets q\left(\vec{u}_{k,t}; {\calR}_{w,k}\right)$
\State Broadcast $\bw_{k,t}$
\EndFor
\State Sample $\zeta$ uniformly from $\{0,1,\ldots,T-1\}$
\State $\widetilde{\bw}_{k+1} \gets \bw_{k,\zeta}$
\EndFor
\State \textbf{Return:} $\widetilde{\bw}_{K+1}$
\end{algorithmic}
\end{algorithm}
\begin{exmp}[URQ]
Consider Definition~\ref{def: Quantization-space}. For any $\bw \in \Conv(\calR)$, define URQ  $q(\bw; \calR) = \bv_i$ with probability $p_i$ such that $\sum\nolimits_i p_i = 1$ and $\sum\nolimits_i p_i \bv_i = \bw$. By the definition, $q$ has the following properties for $\bw \in \Conv(\calR)$:
\newline ~~ -  Unbiasedness: $\mathbb{E}[q(\bw; \calR)] = \bw$, and
\newline ~~ -  Error boundedness: $\|q(\bw; \calR) - \bw\| \leq \max_{i,j} \|\bv_i - \bv_j \|$.
\end{exmp}
Let $b_w$ and $b_g$ be the number of bits used to quantize parameter and gradient vectors in the inner-loop of the quantized SVRG. The number of bits at every (outer-loop) iteration is $64dM+T(b_w+b_g)$, assuming double-precision floating points for the unquantized gradients of dimension $d$ in the outer-loop.
Different from \citep{de2018high}, Algorithm~\ref{alg:QSVRG1} allows for the exchange of low-precision gradients in the inner-loop.
To properly handle this extension, we need to modify the original unquantized SVRG algorithm using a memory unit, as described in the next section.

Abstractly, iterative algorithms search in a space feasible solutions to find the minimizer $\bw^{\star}$. Pseudo-contractive algorithms, which satisfy Assumption \ref{assumption: mubeta}, inherit a nice property that $\|\bw_k - \bw^{\star}\|$ and $\|\bg(\bw_k)\| $ get smaller with $k$. In other words, by every step, we can shrink the feasibility space. Therefore, using the same number of bits, we may reduce the distance among those quantization points, reducing $\left\|q\left(\bw_k\right) - \bw_k \right\|^2$. Under some technical conditions, formally defined in Section~\ref{sec: MainResults}, this error may become sufficiently small for a large $k$, and we may recover full precision of $\bw_k$ with a fixed number of bits. This is possible by an adaptive quantization grid, depicted in Fig.~\ref{fig:Quant} for coordinate $i$ and $j$. Given a proper center for the grid, $\bc_k$, the grid mimics the local geometry of the optimization landscape at iteration $k$ and ensures that $\bw_{k+1}$ (and ultimately $\bw^\star$ as $k \to \infty$) remains inside the grid by proper choice of $[\br_{k}]_i$ for every coordinate $i$.

\section{Main Results}\label{sec: MainResults}
In this section, we present a series of Propositions and Corollaries that characterize the convergence behavior of the quantized SVRG. We show that a naive quantization with few bits and a fixed grid may lead to a suboptimality gap, which can be eliminated by either increasing quantization budget ($b_w$ and $b_g$) or using local geometry of the optimization landscape through our adaptive quantization scheme.

\begin{prop}\label{prop: QSVRG1}
Consider a URQ with fixed quantization grid, i.e., $ \calR_{w,k}$ and $\{\calR_{g_\xi,k}\}_{\xi \in [N]}$ are fixed for all iterations $k$. Let $\alpha_k < 1/6L$ and $T> 1/(\mu \alpha_k (1-6L \alpha_k ))$. Set $\Delta_k := \mathbb{E} \left[ f(\widetilde{\bw}_k)\right] - f(\bw^\star) $, $\beta_{k,t} := \mathbb{E} \left[\left\|q\left(\bu_{k,t}; \calR_{w,k}\right) - \bu_{k,t} \right\|^2 \right]$, and
${\delta}_{k} := \mathbb{E}\left[ \left\|q\left(\bg_\xi \left(\widetilde{\bw}_{k}\right); {\calR}_{g_\xi,k}\right) - \bg_\xi \left(\widetilde{\bw}_{k} \right) \right\|^2\right]$. The iterates of Algorithm~\ref{alg:QSVRG1} satisfy for any $k \in [0, K-1]$
\begin{align}
\Delta_{k+1} - \gamma_k \leq \sigma_k \left(\Delta_k - \gamma_k\right) \:,
\end{align}
where $\sigma_k \in (0,1)$ and
\begin{align*}
\sigma_k = \frac{\frac{1}{\mu T} + 3L\alpha_k^2 }{\alpha_k - 3L\alpha_k^2} \:,
\gamma_k = \dfrac{3 T \alpha_k^2 {\delta}_{k} + \sum_{t = 1}^{T} \beta_{k,t}}{2 T\alpha_k - 12LT\alpha_k^2 - 2/\mu}
\:.
\end{align*}
\end{prop}
Proposition~\ref{prop: QSVRG1} implies that, with the uniform bit allocation in URQ, the quantized SVRG (with fixed grids) converges at a similar rate as the original SVRG until it reaches an ambiguity ball, determined by the combined quantization errors of the gradients (in uplink) and parameters (in downlink), namely $3 T \alpha_k^2 {\delta}_{k} + \sum_{t = 1}^{T} \beta_{k,t}$.

In the following, we show that we can gradually squeeze the quantization grids to constantly reduce $\beta_{k,t}$ and $\delta_{k}$ at every iteration $k$. Thereby, we may maintain linear convergence with a fixed and limited number of bits. To this end, we slightly modify the SVRG algorithm by adding one memory unit at the master node to save  parameter and gradients of the last step. After the inner loop, if the new solution $\widetilde{\bw}_{k}$ has a higher gradient norm, we reject this solution and go to the inner loop by the previous one. Hereafter, we call this slightly modified SVRG algorithm as M-SVRG, and its quantized version by QM-SVRG. Now, one can observe that (details in \citep{Ghadikolaei2019communicationSVRG})
\begin{subequations}
\begin{align}\label{eq: QSVRG-A-rxk}
& \|\widetilde{\bw}_{k+1} - \widetilde{\bw}_{k} \|  \leq  \|\widetilde{\bw}_{k+1} - {\bw}^{\star} \| +  \|\widetilde{\bw}_{k} - {\bw}^{\star} \| \nonumber \\
& \hspace{22mm}\leq \frac{\| \widetilde{\bg}_{k+1} \| + \| \widetilde{\bg}_{k} \| }{\mu} \leq \frac{2 \| \widetilde{\bg}_{k} \| }{\mu} := r_{wk} , \\
\label{eq: QSVRG-A-rgk}
& \left\|\bg_\xi \left(\widetilde{\bw}_{k+1}\right) - \bg_\xi \left(\widetilde{\bw}_{k}\right) \right\| \leq \frac{2 L \| \widetilde{\bg}_{k} \|}{\mu} := r_{g k} \:.
\end{align}
\end{subequations}
These inequalities ensure that $\widetilde{\bw}_{k+1}$ (no matter its true value or quantized version), should be within $r_{wk}$ distance of $\widetilde{\bw}_{k}$. Therefore, we can construct an adaptive grid size to quantize parameter vector at the master node by setting ${\calR}_{w,k} = (\widetilde{\bw}_{k}, r_{wk}, \{b_{wi}\}_i)$. Similarly, we can construct an adaptive grid size to quantize gradient vector at the workers by setting ${\calR}_{g_\xi,k} = (\bg_\xi(\widetilde{\bw}_{k}), r_{gk}, \{b_{gi}\}_i)$.
\begin{prop}\label{prop: QSVRGOurApproach}
Consider a URQ with adaptive quantization grids with ${\calR}_{w,k} = (\widetilde{\bw}_{k}, r_{wk})$ and ${\calR}_{g_\xi,k} = (\bg_\xi(\widetilde{\bw}_k), r_{gk})$  in iteration $k$ where $r_{wk}$ and $r_{gk}$ are defined in \eqref{eq: QSVRG-A-rxk} and \eqref{eq: QSVRG-A-rgk}. Set $\Delta_k := \mathbb{E} \left[ f(\widetilde{\bw}_k)\right] - f(\bw^\star) $. Consider uniform bit allocation to all coordinates and also $b_w = b_g = b$, so $b_{xi}=b_{gj} = b/d$ for all $i,j \in [d]$. Let $\alpha_k < \frac{1}{6L}$ and
\begin{align*}
\frac{b}{d} \geq  \left\lceil \log_2 \left( 1 + \sqrt{\frac{4 L d\left(1+3L^2 \alpha_k^2 \right)}{\mu^2 \alpha_k \left(1 - 6 L \alpha_k \right) }}\right) \right\rceil ~~,~~\text{and} \nonumber \\
T > \frac{1}{\mu \alpha_k \left(1 - 6 L \alpha_k \right) - \frac{4L}{\mu}\left(1+3L^2 \alpha_k^2 \right)\frac{ d}{\left(2^{b/d}-1\right)^2} }
\:.
\end{align*}
The iterates of the modified version of Algorithm~\ref{alg:QSVRG1} satisfy
\begin{align}
\Delta_{k+1} \leq \sigma_k \Delta_k \:,  ~\text{for any}~ k \in [0, K-1],
\end{align}
where $\sigma_k \in (0,1)$ and $\sigma_k = \dfrac{\left( \frac{1}{ T} + 3\mu L\alpha_k^2 + \frac{4L}{\mu}\frac{\left(1+3L^2 \alpha_k^2 \right)d}{\left(2^{b/d}-1\right)^2}  \right)}{\mu \left(\alpha_k - 3L\alpha_k^2\right)}$.
\end{prop}
\begin{cor}\label{cor: QSVRGOurApproach-sigma0}
To ensure a contraction constant of at most $\bar{\sigma}<1$, the minimum number of bits per coordinate and the epoch length should satisfy
\begin{align*}
\frac{b}{d} \geq  \left\lceil \log_2 \left( 1 + \sqrt{\frac{4 L d\left(1+3L^2 \alpha_k^2 \right)}{\mu^2 \alpha_k \left(\bar{\sigma} - 3 L \alpha_k \bar{\sigma} - 3 L \alpha_k\right) }}\right) \right\rceil, ~ \text{and}
\\ T > \frac{1}{\mu \alpha_k \left(\bar{\sigma} - 3 L \alpha_k \bar{\sigma} - 3 L \alpha_k\right) - \left(1+3L^2 \alpha_k^2 \right)\frac{ 4Ld}{\mu\left(2^{b/d}-1\right)^2} } .
\end{align*}
\end{cor}
Proposition~\ref{prop: QSVRGOurApproach} indicates that our proposal recovers linear convergence rate to the optimal solution with a contraction factor of $\sigma_k$. From, Corollary~\ref{cor: QSVRGOurApproach-sigma0}, a sufficient condition on scaling of the number of bits per dimension ($b/d$) is $\log_2(\sqrt{d})$, which is an almost negligible: increasing $d$ from 10 to 1000 may lead to a penalty of 3 additional bits. Moreover, compared to the deterministic approaches, we observe an extra $\log_2(\sqrt{d})$ penalty, which is due to the stochastic gradient terms in the inner loop of Algorithm~\ref{alg:QSVRG1}.

\section{Experimental Results}\label{sec: experiments}
\subsection{Settings}
In this section, we numerically characterize the convergence of our quantized SVRG algorithm on some real-world datasets. For the quantization for each coordinate $i$ of vector $\bv$, we map $[\bv]_i$ onto one of the two closest vertices in the quantization grid with probabilities that are inversely proportional to their distance to $[\bv]_i$~\cite{de2018high}, leading to an unbiased quantization. We set $\lambda = 0.1$ and consider a logistic ridge regression:
\begin{align*}
\min_{\bw} f(\bw) = \frac{1}{N} \sum_{i \in [N]} \ln \left( 1 + e^{-\bw{\tran} \bx_i y_i}\right)  + \lambda \|\bw\|_2^2 \:,
\end{align*}
where each summand is $f_i(\bw)$. Define $\bz_i := \bx_i y_i$. Let $\eig_{\max}(\bX)$ and $\eig_{\min}(\bX)$ denote the largest and smallest eigenvalues of matrix $\bX$. Now, we characterize the geometry of our problem using $L \geq \eig_{\max}\left( \nabla^2 f(\bw) \right)$ and $\mu \leq \eig_{\min}\left( \nabla^2 f(\bw) \right)$ inequalities. We have,
\begin{itemize}
\item Smoothness parameter $L$:
\begin{equation*}
\eig_{\max}\left( \nabla^2 f(\bw) \right) \leq \frac{1}{4N} \sum\nolimits_{i \in [N]} \|\bz_i \|_2^2 + 2 \lambda := L\:.
\end{equation*}
\item Strong conv. parameter:
$\eig_{\min} ( \nabla^2 f(\bw) ) \hspace{-1mm} \geq 2 \lambda := \mu .
$
\end{itemize}

\emph{Datasets:} For the binary classification, we consider from the UCI repository the Individual Household Electric Power Consumption dataset, which has 2,075,259 samples $(\bx_i, y_i)$ of dimension $d=9$. We apply a hard threshold technique on the value of one output to make it a binary class. For the multiclass classification task, we use MNIST dataset, which has 60,000 training samples $(\bx_i, y_i)$ of dimension $d=784$ and 10 classes corresponding to hand-written digits. We applied the one-versus-all technique to solve 10 independent binary classification problems (using logistic ridge regression) and obtain 10 optimal classifiers, $\bw^{(l)}$, one for each digit. Given a test data $\bx$, we choose label $l$ that maximizes $\left(\bw^{(l)} \right){\hspace{-1mm}\tran} \bx$.

\emph{Benchmark Algorithms:} We have implemented SVRG, modified SVRG (M-SVRG), gradient descent (GD)~\citep{Bottou2018SIAM}, stochastic gradient descent (SGD)~\citep{Bottou2018SIAM}, and stochastic average gradient (SAG) \citep{schmidt2017minimizing}.
We should highlight that we count outer-loops of SVRG (also its variants) as the iterations. In terms of bits, following is the number of bits required to run one iteration:
\begin{align*}
& \text{SGD and SAG}  = 128d \:, \quad \text{GD}= 64 d (1+N)\:, \quad \text{and} \\
& \hspace{10mm} \text{SVRG and M-SVRG} = 64dN + 192dT \:,
\end{align*}
where $192d$ corresponds to sending two gradients and one parameter vectors of size $d$ with $N$ workers.

We have also implemented the quantized version of M-SVRG using fixed lattice (QM-SVRG-F) adaptive lattice (QM-SVRG-A) as well as that of benchmark algorithms (GD, SGD, and SAG). We also implement QM-SVRG-F+ and QM-SVRG-A+ in which we quantize both $\bg_\xi(\bw_{k,t-1})$ and $\bg_\xi(\widetilde{\bw}_{k})$ using grid $\calR(g_\xi,k)$ in the inner loop. For the benchmarks, we apply quantizer with a fixed lattice to gradient and parameter vectors, as of the QM-SVRG-F algorithm. Following is the number of bits required to run one iteration:
{\small{
\begin{align*}
& \text{Q-SGD and Q-SAG}= b_w+b_g \:, \quad \text{Q-GD}= b_w + b_g  N \:, \\
& \text{QM-SVRG-F and QM-SVRG-A}= 64dN + 64dT + (b_w+b_g)T ,\\
& \text{QM-SVRG-F+ and QM-SVRG-A+}= 64dN + (b_w+b_g)T \:.
\end{align*}
}}
\emph{Performance Measures:} We have considered the convergence of loss function (zero-order stopping criterion) and the gradient norm (first-order stopping criterion) for the training data, and the F1-score of $\bw_k$ on the test dataset to assess the generalization performance. All the simulation codes are available at ~\citep{Ghadikolaei2019communicationSVRG}.

\subsection{Results and Discussions}
\begin{figure}[t!]
\begin{center}
\begin{minipage}{0.85\columnwidth}
\centerline{\footnotesize\input{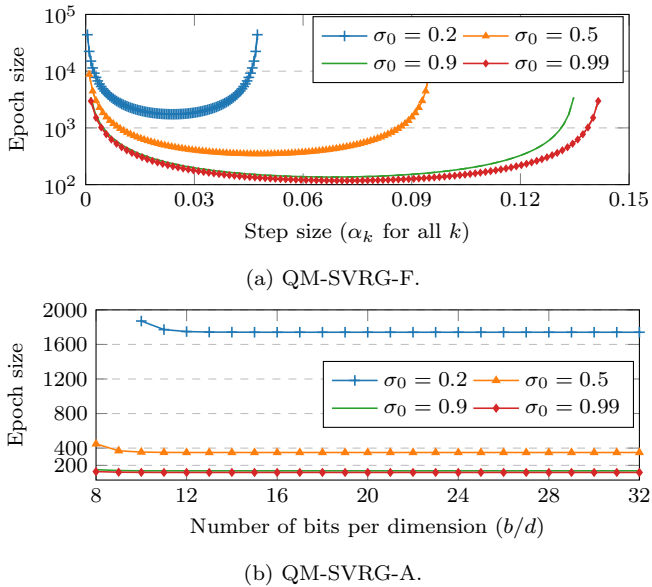}}
\subcaption{QM-SVRG-F.}
\label{subfig: Epoch_size_LB_power_1}
\end{minipage}
\begin{minipage}{0.85\columnwidth}
\centerline{\footnotesize
\begin{tikzpicture}

\definecolor{color0}{rgb}{0.12156862745098,0.466666666666667,0.705882352941177}
\definecolor{color1}{rgb}{1,0.498039215686275,0.0549019607843137}
\definecolor{color2}{rgb}{0.172549019607843,0.627450980392157,0.172549019607843}
\definecolor{color3}{rgb}{0.83921568627451,0.152941176470588,0.156862745098039}

\begin{axis}[%
width=0.95\columnwidth,
height=0.3\columnwidth,
at={(0in,0in)},
scale only axis,
xmin=8, xmax=32,
xtick={0,4,8,12,16,20,24,28,32},
xlabel near ticks,
xlabel={Number of bits per dimension ($b/d$)},
ymin=30, ymax=2000,
ytick={0,200, 400,800,1200,1600,2000},
yticklabel style={/pgf/number format/.cd,%
          scaled y ticks = false,
          set thousands separator={},
          fixed},
grid style={dashed},
ymajorgrids,
yminorgrids,
ylabel near ticks,
ylabel={Epoch size},
axis background/.style={fill=white},
legend columns=2,
legend style={at={(0.99,0.7)}, anchor=north east,legend cell align=left,align=left,draw=black},
]
\addplot [semithick, color0, mark=+]
table {%
10 1871
11 1772
12 1749
13 1743
14 1742
15 1741
16 1741
17 1741
18 1741
19 1741
20 1741
21 1741
22 1741
23 1741
24 1741
25 1741
26 1741
27 1741
28 1741
29 1741
30 1741
31 1741
32 1741
};
\addlegendentry{$\sigma_0 = 0.2$}
\addplot [semithick, color1, mark=triangle*, mark size=1.4pt]
table {%
8 449
9 369
10 354
11 350
12 349
13 349
14 349
15 349
16 349
17 349
18 349
19 349
20 349
21 349
22 349
23 349
24 349
25 349
26 349
27 349
28 349
29 349
30 349
31 349
32 349
};
\addlegendentry{$\sigma_0 = 0.5$}
\addplot [semithick, color2]
table {%
8 150
9 140
10 137
11 137
12 137
13 137
14 137
15 137
16 137
17 137
18 137
19 137
20 137
21 137
22 137
23 137
24 137
25 137
26 137
27 137
28 137
29 137
30 137
31 137
32 137
};
\addlegendentry{$\sigma_0 = 0.9$}
\addplot [semithick, color3, mark=diamond*, mark size=1.3pt]
table {%
8 128
9 121
10 119
11 118
12 118
13 118
14 118
15 118
16 118
17 118
18 118
19 118
20 118
21 118
22 118
23 118
24 118
25 118
26 118
27 118
28 118
29 118
30 118
31 118
32 118
};
\addlegendentry{$\sigma_0 = 0.99$}
\end{axis}

\end{tikzpicture} }
\subcaption{QM-SVRG-A.}
\label{subfig: Epoch_size_LB_power_2}
\end{minipage}

\caption{Sufficient conditions (lower bounds) on the minimum epoch size $T$ against (\subref{subfig: Epoch_size_LB_power_1}) step-size $\alpha_k$ and (\subref{subfig: Epoch_size_LB_power_2}) number of bits per dimension $b/d$ to ensure contraction factor $\bar{\sigma}$ for the Individual Household Electric Power Consumption dataset. In (\subref{subfig: Epoch_size_LB_power_1}), values of QM-SVRG-A with $b/d = 10$ coincide with the corresponding values of QM-SVRG-F, given in this subfigure.}
\label{fig: Epoch_size_LB_power}
\end{center}
\end{figure}
Fig.~\ref{fig: Epoch_size_LB_power} illustrates the theoretical bounds, derived in Section~\ref{sec: MainResults}. The top figure depicts the theoretical lower bound on the epoch size $T$ that ensures a contraction factor of at most $ \bar{\sigma}$. First of all, depending on the target contraction factor and the number of bits per coordinate $b/d$, the feasibility region of $\alpha_k$ varies. For example, to ensure $\bar{\sigma} = 0.2$ (a very good contraction factor), we need a high resolution quantizer (10 bits per coordinate) and very small step-size of $\alpha_k < 0.047$. However, $\bar{\sigma} = 0.9$ is attainable by 8 bits per coordinate and $\alpha_k$ as large as 0.124. The bottom figure shows the same y-axis for the number of bits. Clearly, increasing the number of bits reduces the required $T$ to ensure contraction $\bar{\sigma}$ but in a saturating fashion. In particular, there is no difference in terms of epoch size requirement as well as convergence (as we shall see later on) between $b/d = 15$ and the usual $b/d = 64$. Another point is that the lower the target contraction factor $\bar{\sigma}$, the higher $b/d$ and minimum $T$. Notice that unlike QM-SVRG-F, our QM-SVRG-A algorithm can recover the optimal solution using a finite number of bits per iteration. In the following, we show that the theoretical bounds to ensure convergence are only sufficient conditions and may be very conservative, and we may be able to quantize in practice well beyond those bounds.

\begin{figure}[t!]
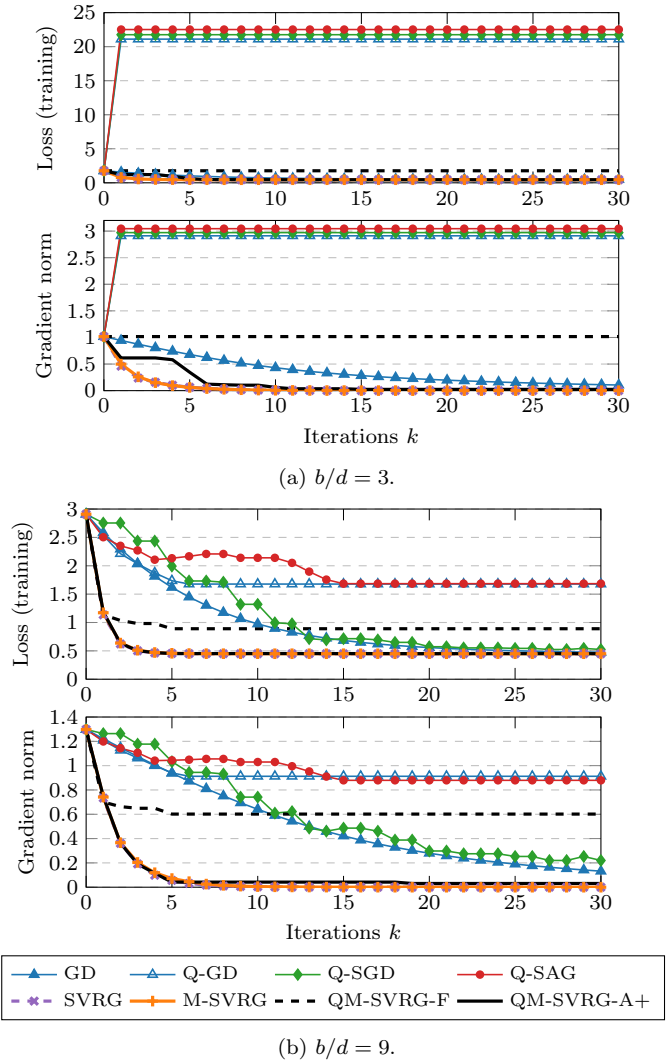

\centering
\begin{minipage}{0.85\columnwidth}
\centerline{\footnotesize\input{P_3_Q_3_withQuant_power_BinaryClassification}}
\subcaption{$b/d = 3$.}
\label{subfig: P_3_Q_3_withQuant_power_BinaryClassification}
\end{minipage}
\begin{minipage}{0.85\columnwidth}
\centerline{\footnotesize\input{P_9_Q_9_withQuant_power_BinaryClassification}}
\subcaption{$b/d = 9$.}
\label{subfig: P_9_Q_9_withQuant_power_BinaryClassification}
\end{minipage}

\caption{Convergence results for Individual Household Electric Power Consumption dataset with $T=8$ and $\alpha_k = 0.2$ for all $k$. Legend Q-x stands for quantized version of algorithm x.}
\label{fig: QuantizationAdaptiveGrid_power}
\end{figure}
Fig.~\ref{fig: QuantizationAdaptiveGrid_power} illustrates the convergence of our performance measures in the presence of quantization.
From Fig.~\ref{subfig: P_3_Q_3_withQuant_power_BinaryClassification}, QM-SVRG-A+ can maintain the convergence to the optimal solution even with a very severe quantization: 3 bits per dimension that is equivalent to 95\% compression in the parameter update and gradient reports. Moreover, this is obtained by $T=8$, well below the values with theoretical guarantees (usually $T> 100$, see Fig.~\ref{fig: Epoch_size_LB_power}). The QM-SVRG-F algorithm (as well as Q-GD, Q-SAG, and Q-SGD), however, cannot keep up with such a severe quantization and maintain convergence, even to a small ambiguity ball. This problem may be addressed only by increasing the number of bits, as we can observe in Fig.~\ref{subfig: P_9_Q_9_withQuant_power_BinaryClassification}. Some algorithms, such as Q-SAG and Q-GD, may be more sensitive and need more quantization bits to recover their convergence. By adding more bits, these algorithms converge at a similar rate of their unquantized version to an ambiguity ball, whose radius determined by the quantization error. Another observation is the noticeable superior performance of the quantized version of M-SVRG, compared to that of other approaches.

\begin{figure}[!t]
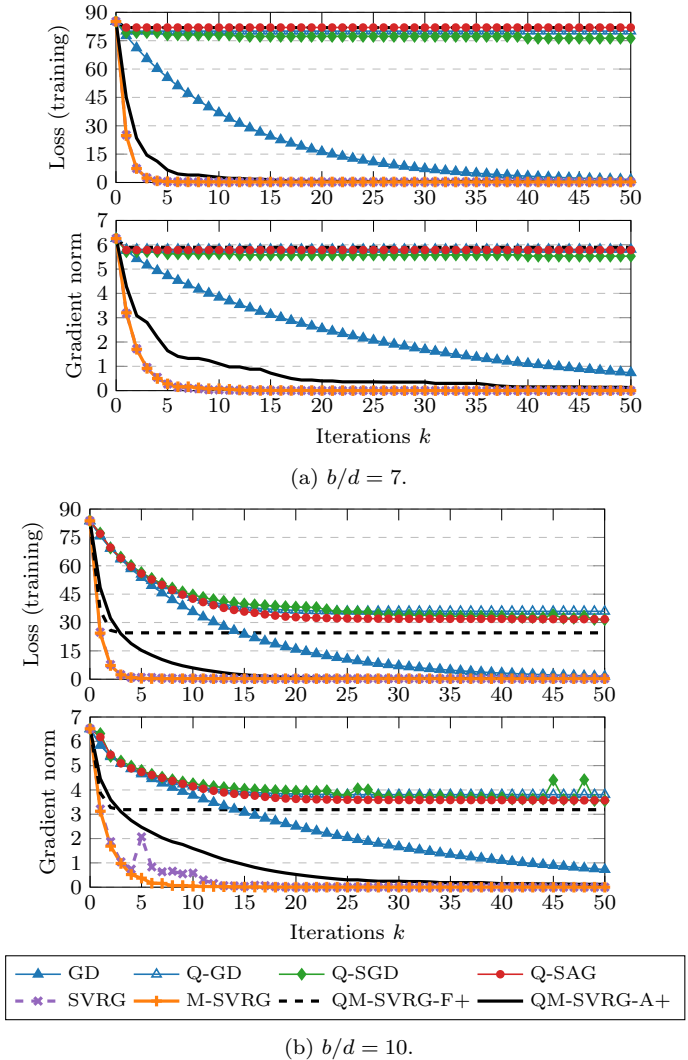

\begin{center}
\begin{minipage}{0.85\columnwidth}
\centerline{\footnotesize\input{P_7_Q_7_withQuant_MNIST_Class9}}
\subcaption{$b/d = 7$.}
\label{subfig: P_7_Q_7_withQuant_MNIST_BinaryClassification}
\end{minipage}
\begin{minipage}{0.85\columnwidth}
\centerline{\footnotesize\input{P_10_Q_10_withQuant_MNIST_Class9}}
\subcaption{$b/d = 10$.}
\label{subfig: P_10_Q_10_withQuant_MNIST_BinaryClassification}
\end{minipage}
\caption{Convergence results for the MNIST dataset (digit 9) with $T=15$ and $(\alpha_k = 0.2)_k$.}
\label{fig: QuantizationAdaptiveGrid_MNIST}
\end{center}
\end{figure}

To asses the impact of higher dimensions and harder learning tasks, we consider a multiclass classification task using the MNIST dataset. Fig.~\ref{fig: QuantizationAdaptiveGrid_MNIST} illustrates the training performance of the classifier for digit 9 with 7 and 10 bits per dimension. Similar to previous cases, the QM-SVRG-A solver can preserve the convergence rate with a severe quantization, thanks to our proposed adaptive grid structure, while other  approaches fail to retrieve the optimal solution.
To better understand the performance of our final solution on all digits, we have reported in Table~\ref{table: QuantizationAdaptiveGrid_MNIST_generalization} the F1-score, averaged over all classes. F1-score is computed assuming digit 9 is the class 1 while all other digits are class -1. M-SVRG slightly outperforms SVRG (not shown in the table) at both $b/d=7$ and $b/d=10$.
We should highlight that we did not try to optimize hyper-parameters to achieve a better F1-score in our experiments. This table, along with Fig.~\ref{fig: QuantizationAdaptiveGrid_MNIST}, indicates a very good performance of our approach compared to the state-of-the-art unquantized and quantized baselines, in terms of training and generalization performance as well as robustness to smaller $b/d$.

\begin{table}[t]
\begin{center}
\captionsetup{width=\linewidth}
\caption{F1-score for the MNIST test dataset with $(\alpha_k=0.2)_k$, $T=15$, and 50 iterations. Table shows the performance of unquantized and quantized algorithms. ``Q-F'' and ``Q-A'' are QM-SVRG-F+ and QM-SVRG-A+.}\label{table: QuantizationAdaptiveGrid_MNIST_generalization}
\renewcommand{\tabcolsep}{1.5mm}
\renewcommand{\arraystretch}{1.1}
{
\begin{tabular}{c|cc|ccccc} \hline \hline
$b/d$ & GD     & M-SVRG & Q-GD  & Q-SGD & Q-SAG & Q-F & Q-A \\ \hline
7                             & 0.775 & 0.841  & 0.127 & 0.101 & 0.130 & 0.139     & 0.806     \\
10                            & 0.780 & 0.841  & 0.248 & 0.402 & 0.168 & 0.280     & 0.838     \\\hline\hline
\end{tabular}
}
\end{center}
\end{table}

\section{Conclusions}\label{sec: conclusions}
We have proposed an adaptive quantization method for SVRG algorithm to maintain its convergence property by communicating only a few bits per iterations. We have theoretically derived the convergence of our novel algorithm and evaluated its performance on a variety of machine learning tasks. Our results indicated that our approach provides a significant compression and is more robust than the quantized versions of other state-of-the-art algorithms for solving distributed optimization problems. Our approach has an immediate use case in the emerging field of machine learning over wireless networks (e.g., internet-of-things) and edge computing.

\bibliographystyle{ifacconf}
\bibliography{refs}

\end{document}